\documentclass{article}
\usepackage{setup}
\usepackage{graphicx,mlspconf}
\usepackage{xcolor}
\usepackage{todonotes}
\usepackage{multicol}
\usepackage{cleveref}
\usepackage{booktabs}
\usepackage{algorithm2e}
\usepackage{mathrsfs}

\copyrightnotice{979-8-3503-2411-2/25/\$31.00 {\copyright}2025 IEEE}

\toappear{2025 IEEE International Workshop on Machine Learning for Signal Processing, Aug.\ 31-- Sep.\ 3, 2025, Istanbul, Turkey}

\title{Continuous-Time Signal Decomposition:\\An Implicit Neural Generalization of PCA and ICA}
\name{%
   Shayan K. Azmoodeh$^{\star}$%
   \qquad Krishna Subramani$^{\star}$%
   \qquad Paris Smaragdis$^{\star}$%
}
\address{%
   $^{\star}$University of Illinois at Urbana-Champaign
}

\begin{document}
\ninept

\maketitle

\begin{abstract}
We generalize the low-rank decomposition problem, such as principal and independent component analysis (PCA, ICA) for continuous-time vector-valued signals and provide a model-agnostic implicit neural signal representation framework to learn numerical approximations to solve the problem. Modeling signals as continuous-time stochastic processes, we unify the approaches to both the PCA and ICA problems in the continuous setting through a contrast function term in the network loss, enforcing the desired statistical properties of the source signals (decorrelation, independence) learned in the decomposition. This extension to a continuous domain allows the application of such decompositions to point clouds and irregularly sampled signals where standard techniques are not applicable. 
\end{abstract}

\section{INTRODUCTION}
\label{sec:intro}

\subsection{Traditional Low Rank Decompositions}
\label{subsec:traditional-prob}

Principal Component Analysis (PCA) and Independent Component Analysis (ICA) are foundational techniques in statistical signal processing and dimensionality reduction \cite{shlensTutorialPrincipalComponent2014, roweisEMAlgorithmsPCA1997, hyvarinenIndependentComponentAnalysis2001}. Both methods aim to recover latent source signals from observed mixtures by identifying a (linear) transformation that reveals underlying structure in the data. PCA achieves this by finding statistically \emph{decorrelated} source components, whereas ICA seeks maximally statistically \emph{independent} components through higher-order statistics \cite{roweisUnifyingReviewLinear1999}. These methods have widespread application in dataset feature generation and blind signal separation.

\begin{figure}[!ht]
%
\begin{minipage}{1.0\linewidth}
    \includegraphics[width=\linewidth]{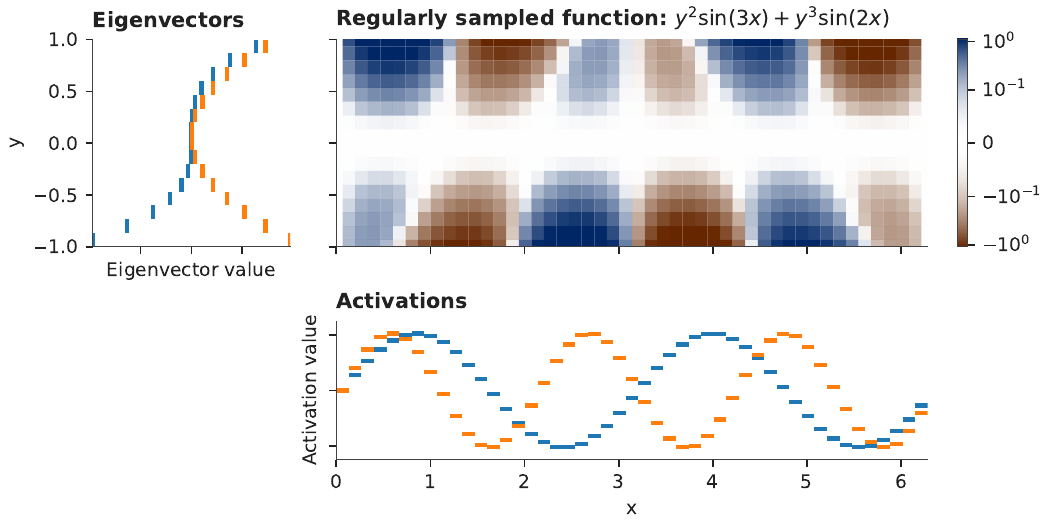}
\end{minipage}
\begin{minipage}{1.0\linewidth}
    \includegraphics[width=\linewidth]{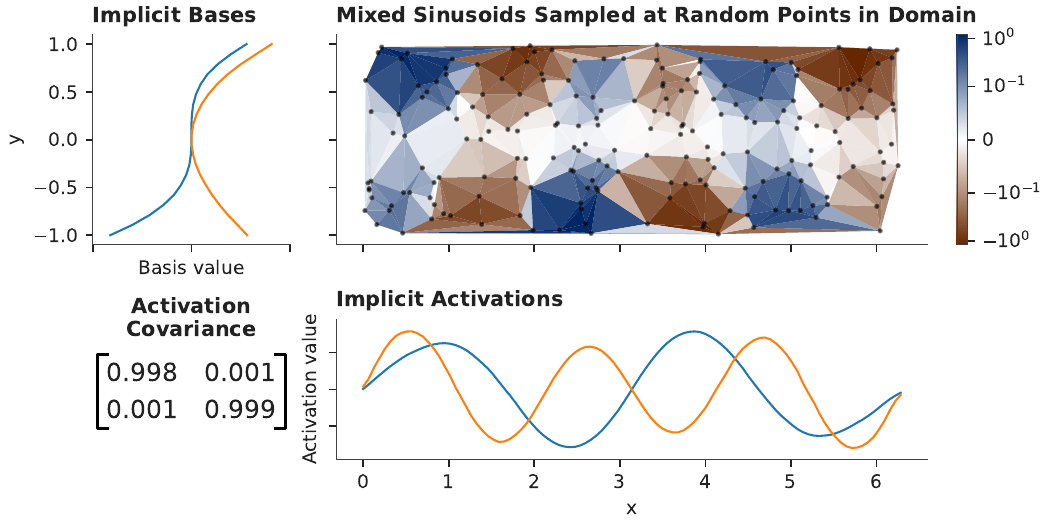}
\end{minipage}
\caption{The plots above show the continuous function $g(x, y) = y^2 \sin(3x) + y^3\sin(2x)$ evaluated at various sample points in $\bR^2$. For the top image, the $x$ and $y$-axes are sampled uniformly giving the output a rigid matrix structure, which we can use to perform PCA in the traditional sense, obtaining discrete vectors with the resulting components.  The bottom plot shows $g$ sampled at irregularly spaced points on its axes; the sample contains the same underlying information, the function $g$, but standard matrix decompositions cannot be applied. The neural implicit PCA formulated here is applied to learn an orthogonalizing decomposition of this irregular sample, extracting the same information as traditional matrix PCA.  Note that the extracted elements are now continuous functions, as opposed to fixed-size vectors which PCA provides.}
\label{fig:wtf}
\end{figure}

Traditional formulations of PCA and ICA operate on finite-dimensional vectors (discretely indexed), and there exist established algorithms to compute the source signal vectors from given datasets \cite{shlensTutorialPrincipalComponent2014, hyvarinenIndependentComponentAnalysis2001}. That is, data is typically assumed to be available in the form of uniformly sampled multivariate time series or tabular datasets (i.e., matrices) similar to the top image of \autoref{fig:wtf}. However, many real-world signals, such as audio, motion capture, or financial time series are more naturally modeled as continuous-time phenomena. In these settings, data may be irregularly sampled or sparsely observed producing datasets in the form of the bottom plot of \autoref{fig:wtf}, making the standard PCA and ICA formulations inadequate or ill-posed.

To bridge this gap, we propose a generalization of the low-rank decomposition problem to continuous-time signals by modeling them as a sample path of continuous-time stochastic processes, allowing for flexible representations of data without requiring uniform sampling or fixed-length sequences. Leveraging implicit neural representations, we learn smooth signal decompositions that enforce the desired statistical properties of latent sources through contrastive loss functions, unifying PCA and ICA under a single framework.

\subsection{The Continuous-Time Problem}
\label{subsec:contin-problem}

All random variables are defined as functions on the probability space $(\Omega, \mathcal{F}, \mathbb{P})$, where $\Omega$ is the abstract event space, $\mathcal{F}$ defines all subsets of $\Omega$ which have a measurable probability, and $\mathbb{P}$ is the probability measure. We can now present the PCA and ICA problems in their full generality for continuous-time signals. Let $\mathbf{S}_t = \Bigl[ S^{(1)}_t , \ldots, S^{(k)}_t \Bigr]^{\top} \; \in \; \bR^{k}$, $t \in I$ for some closed, bounded interval $I$ in the positive reals, be the vector-valued stochastic process with continuous sample paths representing the (unobserved) source signals. Note this implies that each component of a sample path of $\mathbf{S}_t$ is a continuous real-valued function on $I$ (a function of $t$).
We make the following assumptions on the source signals as in PCA and ICA \cite{shlensTutorialPrincipalComponent2014, amariNewLearningAlgorithm1995, cardosoSourceSeparationUsing1989}.

\begin{assumption}\label{assump:xt-stationary}
    For each $i = 1, \ldots, k$, $S_t^{(i)}$ forms a stationary, square integrable process, i.e., each $S_t^{(i)}$ has finite variance.
\end{assumption}

The stationary assumption allows us to remove time dependence on any expectations on elements on the process since such statistics remain the same across time; we denote the process in such cases where the result is time agnostic simply by $\mathbf{S}$. We add an additional assumption of finite moments of all orders when solving the ICA problem to allow for the use of these higher order moments to approximate statistical independence among learned source components. The expectation of a random variable $X$ is denoted by $\E{X} \triangleq \int_{\Omega} X(\omega) \,\dd{P}(\omega)$.

\begin{assumption}[for ICA]\label{assump:xt-finite-moments-ica}
    \[
        \forall m \in \bN,\, \forall i \in \set{1, \ldots, k},\, \E{\bigl\lvert S^{(i)} \bigr\rvert^m} < \infty.
    \]
\end{assumption}

Let $\mathcal{X}$ be a separable Banach vector space (providing existence guarantees for the observed process) denoting the state space of each variable in the observed signal process $(X_t)_{t \in I}$ and let $\mathbf{T}: \bR^k \rightarrow \mathcal{X}$ be an (unknown) linear operator. \emph{Given only the observed signal  $X_t = \mathbf{T}\cdot \mathbf{S}_t$, we wish to recover $\mathbf{S}_t$ for each $t \in I$ and the operator $\mathbf{T}$.} In the case of PCA, we wish to find $\mathbf{S}_t$ such that the components of the source signal are decorrelated\footnote{Note that strictly speaking, to perform PCA we also have a requirement that the basis functions are also mutually orthogonal. This is an extra constraint that we can easily add in our proposed method later on, however for the sake of simplicity and generality we only consider PCA to simply be a decorrelating transform in this paper.}, i.e., $\forall t \in I$, for $i, j \in \set{1, \ldots, k}$,
\begin{align}\label{eqn:decorr-S}
    \E{S^{(i)}_t S^{(j)}_t} =
    \begin{cases}
        0 & i \ne j\\
        1 & i = j.
    \end{cases}
\end{align}
For ICA, we require the stronger condition of maximal statistical independence among the components of $\mathbf{S}_t$ at each $t$.

The function presented in \autoref{fig:wtf} can be understood under this framework by equating the time-domain of $X_t$ with the $x$-domain of $g$, i.e., $X_t = g(t, \cdot)$ a continuous function of $y$. Then at each $t$, the process realizes a continuous function of $y$, so we have our state space $\mathcal{X}$ being the space of continuous functions on $[-1, 1]$. This process can be decomposed into the shown source signals and basis functions. The orange and blue sinusoidal source signals correspond to $S_t^{(1)} = \sin(3t)$ and $S_t^{(2)} = \sin(2t)$ respectively, again equating $t$ here with $x$ in the figure. The shown basis functions, $f_1(y) = y^2$ (orange) and $f_2(y) = y^3$ (blue), define the operator $\mathbf{T}$ by $\mathbf{T}\cdot\mathbf{S}_t = f_1 S_t^{(1)} + f_2 S_t^{(2)}$.

This continuous-domain problem parallels the traditional discrete PCA/ICA. Namely, the source signal process is analogous to the unobserved source signal, the operator $\mathbf{T}$ parallels the mixing matrix, and of course the observed signal process parallels the observed/mixed signal (refer \autoref{table:comp-cont-discr}). With this, one can see that the standard discrete PCA/ICA mixture problems are a subset of this general setup. Letting $\mathcal{X} = \bR^d$ ($d \ge k$), we can let $\mathbf{T}: \bR^d \rightarrow \bR^k$ be a matrix in $\bR^{k \times d}$. Given the discrete, stationary source signal process $\mathbf{S}_{t_1}, \mathbf{S}_{t_2}, \ldots$ for $t_1, t_2, \ldots, \in I$ (given any discrete indices we can renumber them to be in $I$), we can extend it to a continuous time process $(\mathbf{S}_t)_{t \in I}$ by letting $\mathbf{S}_{t} = \mathbf{S}_{t_i}$ almost surely for $t  \in [t_i, t_{i+1})$ (i.e., fill in the gaps in the process from $t_i$ to $t_{i+1}$ with $\mathbf{S}_{t_i}$). Then, for $t \in [t_i, t_{i+1})$, $\mathbf{S}_t$ behaves statistically like the single random vector $\mathbf{S}_{t_{i}}$ (see Remark 4.2 \cite{schellNonlinearIndependentComponent2023}), and thus $X_t = \mathbf{T}\cdot\mathbf{S}_t$ behaves like $X_{t_i} = \mathbf{T}\cdot\mathbf{S}_{t_i}$. Thus, by solving this more general problem for $(\mathbf{S}_t)_{t \in I}$ and $\mathbf{T}$, we can sample this process at $t = t_{1}, t_2, \ldots$ to obtain a solution $\mathbf{S}_{t_1}, \mathbf{S}_{t_2}, \ldots$ along with the matrix $\mathbf{T}$ for the original discrete signal.

\begin{table}[ht]
\centering
\caption{Comparison of Continuous and Discrete Problems}
\begin{tabular}{c|cc}
    \toprule
    & \textbf{Continuous} & \textbf{Discrete} \\
    \midrule
    \textbf{Observed Data} & $X_t \in \mathcal{X}$ & $\mathbf{x}_t \in \mathbb{R}^d$ \\
    \textbf{Latent Sources} & $\mathbf{S}_t \in \mathbb{R}^k$ & $\mathbf{s}_t \in \mathbb{R}^k$ \\
    \textbf{Mixing Operator} & $\mathbf{T}: \mathbb{R}^k \to \mathcal{X}$ & $\mathbf{T} \in \mathbb{R}^{d \times k}$ \\
    \textbf{Time Index} & $t \in \mathbb{R}$ & $t \in \{1, 2, \dots, p\}$
\end{tabular}
\label{table:comp-cont-discr}
\end{table}

We model a finite, discrete dataset $\mathcal{D} \subseteq \mathcal{X}$ with $N$ points under this problem setup as a discrete sampling in process time of a sample path of the continuous underlying mixture process $(X_t)_{t \in I}$. Note we make a distinction between the sample paths of the process (a realization of a value by each random variable in the process) and the sampling of the process in time (observing a continuous sample path at discrete time points). The term ``time'' here refers to the \emph{process time} of the underlying stochastic process used to model the dataset, not necessarily a time dimension in the original signal or data; this process time models the index domain of points in the dataset, i.e., the $x$-domain of the function sampled in \autoref{fig:wtf}. Thus we can write the dataset as $\mathcal{D} = \set{X_{t_1}(\omega), X_{t_2}(\omega), \ldots, X_{t_N}(\omega)}$ for $t_1, \ldots, t_N \in I$ and some fixed $\omega \in \Omega$. Treating the dataset as a sample path of a stochastic process limits us to having just one sample from each random variable in the process. The stationary assumption \ref{assump:xt-stationary} enables computing expectations from the given observed data.

The idea of decomposing into functional components rather than discrete vectors has been previously explored in the framework of functional principal component analysis (fPCA) \cite{dauxoisAsymptoticTheoryPrincipal1982, yao2005functional}, as well as in modern autoencoder-based approaches \cite{zhong2023nonlinear}. fPCA models data as multiple realizations (sample paths) of a continuous-time stochastic process and extracts continuous basis functions that capture the principal modes of variation across these paths. As such, it assumes access to a finite population of discretely indexed observations with the same sample domain, i.e., each data point is a function of one variable. In contrast, our formulation generalizes this setup in several ways. First, it accommodates stochastic processes/signals taking values in arbitrary vector spaces, including function spaces, allowing for greater modeling flexibility. Second, it removes the restriction that data must be discretely indexed, enabling analysis of datasets with irregular or continuous sampling across both dimensions (see \autoref{fig:wtf} for example). However, this generalization comes with the trade-off of operating on a single observed sample path of the underlying process, whose implications were discussed above. Crucially, our framework also unifies the PCA and ICA problems under a single decomposition formulation. While the general theory of vector-valued random functions in \cite{dauxoisAsymptoticTheoryPrincipal1982} encompasses our setting in principle, it does not yield a practical or natural computational procedure for solving the decomposition problem as described.

\section{LEARNING IMPLICIT DECOMPOSITIONS}
\label{sec:learn-decomp}

\newcommand{\Lreconstr}{
    \mathscr{L}_{\text{reconstr}}
}

We provide a model-agnostic \emph{unifying} neural network-based approach using implicit neural signal representations to solve the problem as stated in \autoref{subsec:contin-problem} numerically (as an optimization problem) for both PCA and ICA in the general case where $\mathcal{X} = C([a, b])$, the space of continuous functions $f: [a, b] \rightarrow \bR$, i.e., each element in the observed process/signal is a continuous real-valued function, and given a dataset $\mathcal{D}$, where the $i$-th data point is a discrete sampling of the function realized at the observed $i$-th time step (note we do not impose all elements in the dataset to be vectors of the same size, that is, each data point can be a different sized vector resulting from sampling the underlying function at varying points). Without loss of generality we assume $\mathcal{X} = C([0, 1])$ and the time domain of the continuous process is $I = [0, 1]$, since we can apply a scaling/translation bijection between $[0, 1]$ and any interval on the inputs to the function. Since each realization of $X_t$ is a function, notationally we write $X_t(\xi)$ to denote the scalar random variable obtained by evaluating the realized function at the point $\xi$.
 
The function space $C([0, 1])$ can be considered as an infinite dimensional vector space, thus we can consider this problem in a similar light to the discrete case: we wish to find a set of basis functions $f_1, \ldots, f_k \in C([0, 1])$ (i.e., the ``columns'' of the operator ``matrix'' $\mathbf{T}$) and corresponding activation/weight random variables $H_1, \ldots, H_k$ that realize values in $C([0, 1])$ (functions of process time, each representing a component of $\mathbf{S}_t$), that output an activation for each point in time, to form the best ``rank-$k$'' reconstruction of $X_t$ as a linear combination of the basis functions,
\begin{align}\label{eqn:lin-reconstr}
    X_t(\cdot) \approx \mathbf{T}(\cdot)\cdot\mathbf{S}_t = \sum_{n = 1}^{k} H_n(t)\, f_n(\cdot).
\end{align}
If we let $k \rightarrow \infty$, we end-up with the Karhunen-Lo\`eve transform for vector-valued processes \cite{dauxoisAsymptoticTheoryPrincipal1982,loeve1946functions,Levy2008}. We denote the operator $\mathbf{T}$ in familiar matrix notation, $\mathbf{T} = \bigl[ f_1 ~ \cdots ~  f_k\bigr]$ to emphasize the similarities to the discrete PCA/ICA problems (then evaluating $\mathbf{T}(\xi)$ can be thought as accessing the $\xi$-th ``row'' of the ``matrix''). Similarly, we use the term ``rank-$k$'' here to further highlight the connection to the discrete case. The notion of rank here arises when considering a discrete sampling of $\mathbf{S}$ across $t$ to form a matrix and a sampling of the functions comprising the columns of $\mathbf{T}$ to get a matrix approximation of $\mathbf{T}$. The matrix given by this discretization of $\mathbf{T}\cdot\mathbf{S}$ is a rank-$k$ approximation of the matrix formed by a discrete sampling of the observed process $X$ across both time and function input dimensions.

The idea of using neural networks to learn implicit functions has been explored in computer vision and graphics \cite{tancikFourierFeaturesLet2020,sitzmann2020implicit,10.1145/3503250}. They have also been recently explored in the context of non-negative matrix factorization of irregularly sampled time-frequency representations for audio \cite{subramani2024rethinking}. With these insights, we can model each of the $f_n$ and $H_n$ functions by neural networks $\hat{f}_n$ and $\hat{H}_n$ to get estimates of the source signal process $\hat{\mathbf{S}}_t = \Bigl[\hat{H}_1(t), \ldots, \hat{H}_k(t) \Bigr]^\top$ and $\hat{\mathbf{T}} = \bigl[\hat{f}_1 ~ \cdots ~ \hat{f}_k \bigr]$. 
The low-rank approximation of the observed signal is then given by
\begin{align}
    \hat{X}_t(\cdot) = \hat{\mathbf{T}}(\cdot) \cdot\hat{\mathbf{S}}_t = \sum_{n = 1}^{k} \hat{H}_n(t) \hat{f}_n(\cdot).
\end{align}
In practice, we model each of these functions as a standard feedforward network with three hidden layers and PReLU activations between each hidden layer (no activation applied to the output of the final layer). Fourier positional encodings with frequencies randomly sampled from a zero-mean normal distribution (variance is a tunable hyperparameter) are applied to the one-dimensional inputs to the networks to improve training \cite{tancikFourierFeaturesLet2020} \footnote{Code https://github.com/Shkev/implicit-signal-decomps}. We train these networks to minimize a reconstruction loss $\Lreconstr$. Additionally, we add a loss term $\phi: \bR^k \rightarrow \bR_{+}$ acting on the estimated source signal $\hat{\mathbf{S}}_t$ to enforce the statistical properties we desire from the source signal $\hat{\mathbf{S}}_t$; this term is also referred to as a \emph{contrast function} \cite{hyvarinenIndependentComponentAnalysis2001}. In practice we use the mean-squared error as the reconstruction loss,
\begin{align}\label{eqn:reconstr-loss-mse}
    \Lreconstr\bigl(\hat{\mathbf{T}}, \hat{\mathbf{S}} \bigr) &\coloneqq \E{\norm{X - \hat{X}}_{L^2([0,1])}^2} \nonumber\\
    &= \E{\int_{[0, 1]} \Bigl(X(\xi) - \hat{X}(\xi) \Bigr)^2 \,\dd{\xi}},
\end{align}
where the inner integral can be estimated using a Monte Carlo approximation by evaluating $X_t(\xi) - \hat{X}_t(\xi)$ at points $\xi \in [0, 1]$ available in the dataset, and the expectation can be estimated by averaging the approximation of the inner integral at the various time points from the observed dataset (since we assume the source signal components are stationary processes).
The full loss function can be formulated as,
\begin{align}\label{eqn:loss-fnc-full}
    \mathscr{L}\bigl(\hat{\mathbf{T}}, \hat{\mathbf{S}} \bigr) = \Lreconstr\bigl( \hat{\mathbf{T}}, \hat{\mathbf{S}} \bigr) + \beta\, \phi\bigl(\hat{\mathbf{S}} \bigr),
\end{align}
where $\beta$ is a tunable hyperparameter to control the relative weight of the contrast function in learning the decomposition. Training the networks to minimize these functions gives functions that numerically approximate the solution to the problem presented in \autoref{subsec:contin-problem}, recovering the source signal process and the mixing operator. The networks for the bases and activations are learned using gradient descent as shown in Algorithm \ref{alg:implicit-pca-train}.
Computing the loss functions during the training process to learn the implicit functional solutions does not require evenly spaced points as each $\hat{X}_t(\xi)$ can be computed at any time and function input point $(t, \xi)$ by evaluating $\hat{\mathbf{T}}(\xi)\cdot\hat{\mathbf{S}}_t$ using the partially learned networks to compare to the given dataset $X_t(\xi)$ and compute the reconstruction loss and update the network weights. The computation of $\phi$ can also be done using \emph{any} random sampling of time points to estimate expectations since it has no dependence on the input data. We only expect the inputs to be available to us in the form of tuples $\mathcal{D} = \set{(t_i, \xi_i, x_i)}_{i = 1}^{N}$ which can be regularly or irregularly sampled \cite{subramaniPointCloudAudio2021}.


In the case where the state space of the observed signal is finite dimensional, i.e., $\mathcal{X} = \bR^d$ and thus each $\mathbf{X}_t \in \bR^d$, the setup can be modified by setting each $f_n \in \bR^d$ a vector and modifying the reconstruction loss \pref{eqn:reconstr-loss-mse} to be a sum across the $d$ entries in each $\mathbf{X}_t$ in place of an integral.


\begin{algorithm}[ht] \label{alg:implicit-pca-train}
    \caption{$\texttt{NeuralDecomp}\paren{\mathcal{D}, \eta, E}$}
    \DontPrintSemicolon
    \SetKwProg{Define}{Def}{:}{}
    \SetKwInOut{Input}{Input}
    \SetKwInOut{Output}{Output}

    \Input{ Data samples $\mathcal{D} = \set{(t_i, \xi_i, x_i)}_{i = 1}^{N}$\\
            $\eta$: learning rate\\
            $E$: number of epochs}
    \Output{ Neural bases $\hat{f}_1, \ldots, \hat{f}_k$, and source signals $\hat{H}_1, \ldots, \hat{H}_k$}

    \BlankLine

    $\theta \gets$ neural network parameters for $\hat{f}_1, \ldots, \hat{f}_k, \hat{H}_1, \ldots, \hat{H}_k$\;
    \For{$i \gets 1$ \KwTo $E$}{
        \For{$j \gets 1$ \KwTo $N$}{
            $\hat{x}_j \gets \sum_{n = 1}^{k} \hat{H}_n(t_j) \hat{f}_{n}(\xi_j)$\;
            $\hat{\mathbf{T}} \gets \Bigl[\hat{f}_1(\xi_j) ~ \ldots ~ \hat{f}_k(\xi_j) \Bigr]$\;
            $\hat{\mathbf{S}}_{t} \gets \Bigl[\hat{H}_1(t_j), \ldots, \hat{H}_k(t_j) \Bigr]^{\top}$\;
            $\mathcal{L} \gets \mathscr{L}\bigl(\hat{\mathbf{T}}, \hat{\mathbf{S}}_{t} \bigl)$\;
            $\theta \gets \theta - \eta \grad \mathcal{L}$\;
        }
    }
    \Return{$\hat{f}_1, \ldots, \hat{f}_k, \hat{H}_{1}, \ldots, \hat{H}_k$}\;
\end{algorithm}


\subsection{Statistical Contrast Functions}
\label{subsec:contrast-loss}

The contrast function $\phi$ assigns a measure of a statistical property between the entries of a random vector. In the case of PCA, we would like to enforce decorrelated components across each $\mathbf{S}_t$ in the source signal. This leads to the following contrast function for learning the PCA solution,
\begin{align}\label{eqn:pca-contrast-fnc}
    \phi_{\text{PCA}}(\mathbf{S}) = \norm{\E{\bigl(\mathbf{S} - \bvec{\mu}\bigr)\bigl(\mathbf{S} - \bvec{\mu}\bigr)^{\top}} - \mathbf{\Lambda}},
\end{align}
(where $\mathbf{\Lambda}$ is an $k \times k$ diagonal matrix and $\bvec{\mu} \coloneqq \E{\mathbf{S}}$) which has a minimum when the entries in the source signal are decorrelated.

For solving the ICA problem, measuring statistical independence between the components of $\mathbf{S}$ is notoriously difficult as it involves all higher order cross-cumulants of the considered variables, not just those of order 2 as in decorrelation. Following the work of \cite{hyvarinenIndependentComponentAnalysis2001, cichockiRobustLearningAlgorithm1994, shun-ichiamariBlindSourceSeparationsemiparametric1997}, we estimate statistical independence of components of $\mathbf{S}$ through a non-linear decorrelation criterion of the form,
\begin{align}\label{eqn:ica-constrast-fnc}
    \phi_{\text{ICA}}(\mathbf{S}) = \norm{\E{\bigl( \varphi(\mathbf{S}) - \tilde{\bvec{\mu}} \bigr) \bigl( \mathbf{S} - \bvec{\mu} \bigr)^{\top}} - \mathbf{\Lambda}},
\end{align}
(where $\tilde{\bvec{\mu}} \coloneqq \E{\varphi(\mathbf{S})}$) for some arbitrary non-linear activation function $\varphi: \bR \rightarrow \bR$ applied element-wise to the vector $\mathbf{S}$. Minimizing $\phi_{\text{ICA}}$ provides a heuristic to bring the components of $\mathbf{S}$ closer to being statistically independent. We note, however, that this criterion is quite general as most independence criterion used to measure independence for ICA can be reduced to this form based on the choice of activation $\varphi$ \cite{hyvarinenIndependentComponentAnalysis2001}. Moreover, \cite{shun-ichiamariBlindSourceSeparationsemiparametric1997} show that the form of $\phi_{\text{ICA}}$ is as general as possible among other non-linear decorrelation criterion. Thus, the function $\varphi$ can be treated as a hyperparameter that can be tuned to better learn the distribution of the underlying source signal. In practice, we find $\varphi(y) = \tanh(y)$ and (for both PCA and ICA) $\mathbf{\Lambda} = \mathbf{I}_n$ works well.

The contrast function unifies the solving of the PCA and ICA problem. By selecting the appropriate contrast function and training the network to minimize the resulting loss function \pref{eqn:loss-fnc-full}, we learn the desired source signals and mixing operator. Once again, the expectations in the functions above can be approximated by averaging the result across multiple randomly sampled time points.

\begin{figure}[!t]
\centering
\begin{minipage}[c]{\linewidth}
    \centering
    \includegraphics[width=0.9\linewidth]{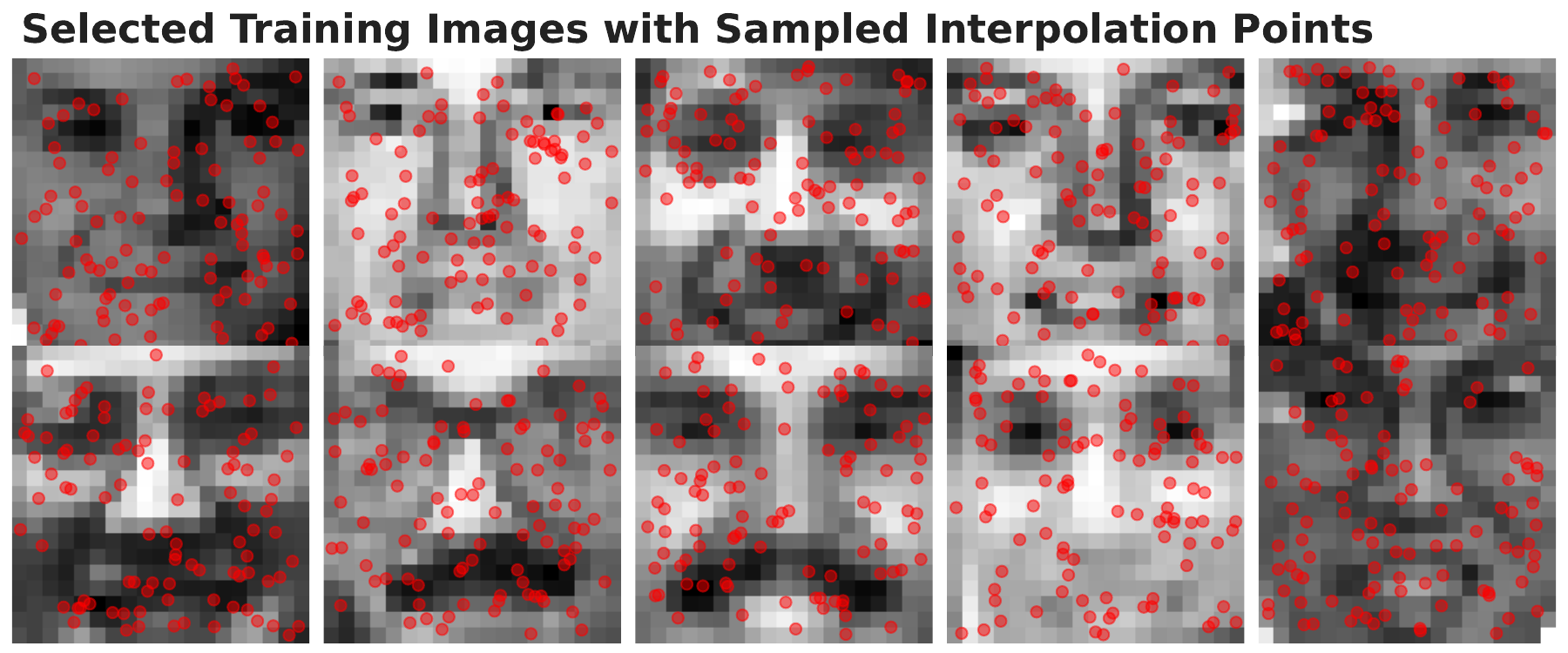}
\end{minipage}\\
\begin{minipage}[c]{\linewidth}
    \centering
    \begin{minipage}[c]{0.35\linewidth}
        \centering
        \includegraphics[width=\linewidth]{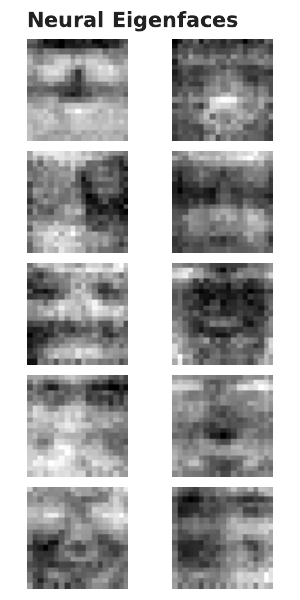}
    \end{minipage}%
    \hfill
    \begin{minipage}[c]{0.65\linewidth}
        \centering
        \includegraphics[width=0.9\linewidth]{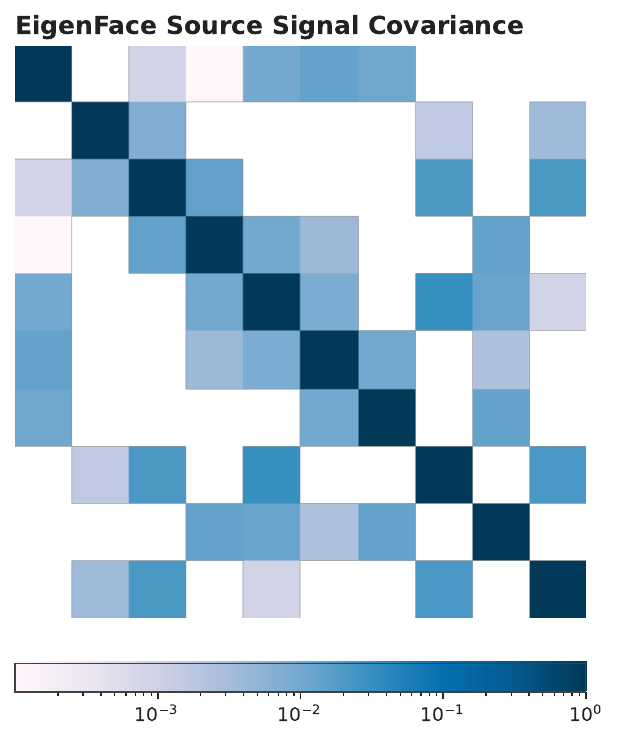}
    \end{minipage}
\end{minipage}%
\caption{
Eigenfaces recovered by an implicit neural PCA network. Input faces, a sample of which is shown in the top panel, were sampled randomly as the points shown by the red dots, and only these values and their real-valued indices were used to perform the analysis. The learned bases, rendered as images, are show on the left and as expected resemble closely the well-known eigenfaces.  The activations covariance is also shown on the bottom right.  Note that the covariance plot is on a log scale, and displays a significant amount of decorrelation.
}
\label{fig:eigface}
\end{figure}

\begin{figure*}
\centering
\includegraphics[width=\linewidth]{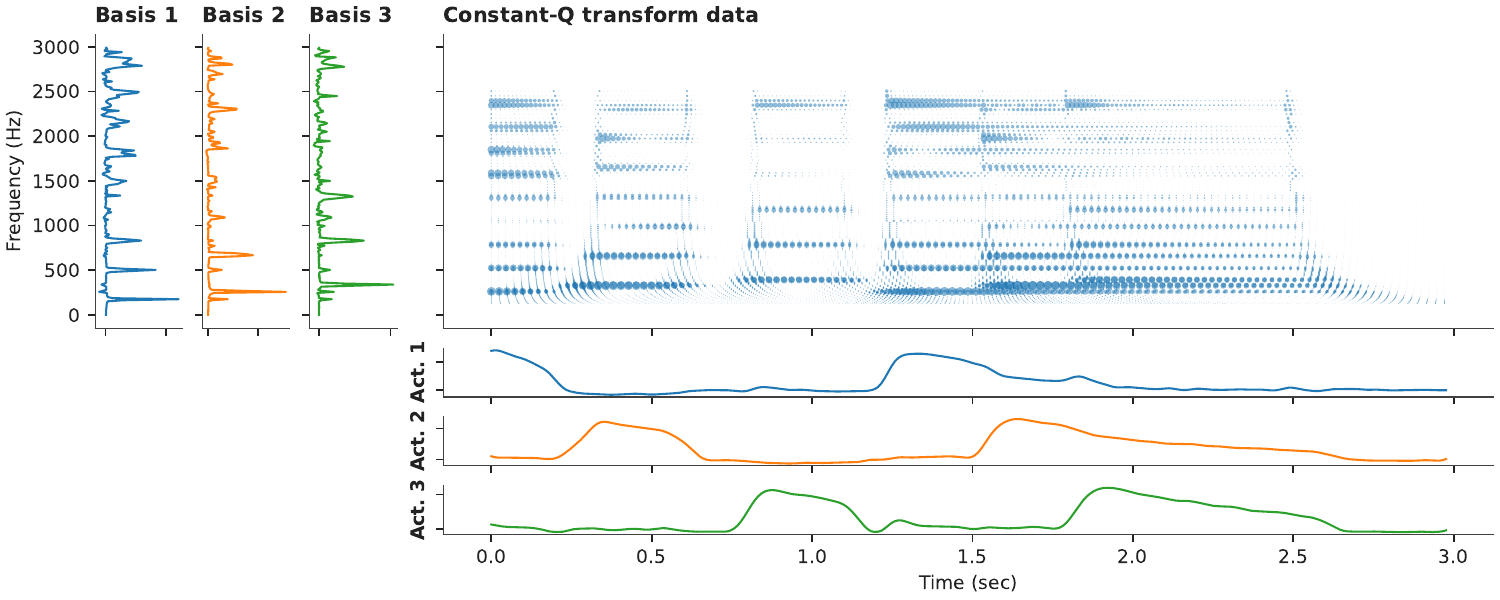}
\caption{The constant-Q transform spectrogram of audio clip of a sequence of three notes played in isolation one after the other, and then again in sequence but this time while overlapping, is shown above as a scatter plot. Note that the frequency and time spacing is variable, and the transform's data cannot be represented as a matrix.  A traditional approach to separating these notes/components is the discrete ICA, relying on their independence of activation, however this is not possible here due to the irregular sampling of times and frequencies across a continuous domain. Using our neural implicit ICA we are able to identify the spectra and activations of these three components as shown above. Note that the ICA bases represent the constant-Q spectra of each of the three notes, and that their corresponding activations show where in time these notes are active.}
\label{fig:audio-cqt}
\end{figure*}

\section{EXPERIMENTS}
In this section we demonstrate the potential of this approach using two examples based on real data.  First, we examine the case of extracting face features via PCA (eigenfaces), but from irregularly sampled images.  We subsequently provide an example of extracting sources from an irregularly sampled time-frequency representation, a constant-Q transform, where again traditional ICA would not be directly applicable.

\subsection{PCA of Faces Data}

The neural signal decomposition algorithm presented in \autoref{sec:learn-decomp} easily extends beyond two-dimensional data to datasets with multidimensional entries, i.e., image datasets where each entry is a 2D image. In the case of a dataset with $n$-dimensional data points, we can simply set the state space of the observed signal process $\mathcal{X} = C([0, 1]^n)$ (as presented in \autoref{subsec:contin-problem}), the space of continuous functions on the $n$-dimensional unit square in $\Rn$. The neural network setup presented in \autoref{sec:learn-decomp} to solve this problem is modified to allow the basis functions to accept $n$ inputs as opposed to one; in the language of the previous section, we require the basis function inputs $\xi$ to be in $\bR^n$. For an image dataset, we model each image as a function of two inputs, the $x$ and $y$ coordinates of each pixel, and the index of each image in the dataset is equated with process time. That is, $X_{t}(x, y)$ models pixel with coordinates $(x, y)$ in the $t$-th image in the dataset. Since the images form a discrete dataset with no continuous structure between them that can be meaningfully interpolated, the activations are modeled as matrices (i.e., a discrete set of vectors as opposed to continuous functions of the data index, so the $t$ domain is discrete); this resulted in faster training and the model learning more visually appealing components. We use the MIT CBCL Face Dataset \#1 \cite{cbcl} (100 faces) for our experiments.

To evaluate our method on face image data, we first applied bilinear grid interpolation to each image to obtain a continuous approximation in two dimensions. This interpolant function was then randomly sampled at various non-lattice points on each image and these sampled pixels formed a collection of irregularly sampled tuples consisting of the image index, the $x$ and $y$ coordinates of the a pixel, and the pixel's magnitude, $(t, x, y, m)$. The implicit neural PCA model is trained on these points to recover the 10 eigenfaces (the learned basis functions) that provide the best rank-10 reconstruction of the original faces in the dataset. Each learned eigenface (see \autoref{fig:eigface}) highlights a different set of facial features such as cheekbones, eyebrows, highlights around the nose, that serve as building blocks of the original facial images. The contrastive loss function in the model enables the model to learn these basis images such that the corresponding source signals (each basis's contribution to each image in the dataset) is decorrelated/orthogonal, as expected from the traditional PCA. The covariance matrix of the source signals of the learned representation is shown in \autoref{fig:eigface} with colors shown on a log-scale. Note that most covariances off the diagonal have magnitude on the order of $10^{-3}$ or less. The shown neural EigenFaces explain $\approx 79\%$ of the variance in the dataset, compared to $\approx 85\%$ with the traditional PCA.

\subsection{Decomposing Irregularly Sampled Spectrograms}
\label{subsec:point-cloud}

We consider the problem of extracting audio components from the frequency domain representation of an audio signal resulting from the constant-Q transform (CQT) \cite{brownCalculationConstantSpectral1991}, where the output spectrogram contains data at unevenly spaced frequencies and time intervals (similar to a wavelet transform), making it unsuitable to represent in a traditional rigid matrix structure. Such a transform has larger frequency resolution in the lower frequencies (i.e., larger window size) with this window size decreasing as the frequencies increase. We use a variant of the CQT presented in \cite{velascoCONSTRUCTINGINVERTIBLECONSTANTQ2011} that has non-uniform spacing of points in the time axis as well. In performing such a transform on a signal, we obtain a magnitude output for each of a set of time/frequency pairs. This lends itself to 3D representation where each data point can be represented as a time, frequency, magnitude $(t, \xi, m)$ tuple. Following the framework described above, the CQT gives us a discrete sampling of the observed mixture process $X_t(\xi) = m$, which we use to find continuous source signals $\mathbf{S}_t$ and basis functions $f_1(\xi), \ldots, f_k(\xi)$ to form the mixing operator $\mathbf{T}$ using the methods of \autoref{sec:learn-decomp}.

The CQT spectrogram of an audio clip consisting of three notes is shown in \autoref{fig:audio-cqt}. In that clip, the three notes are played in  isolation and in sequence, and then again in sequence but with significant overlap. Traditional approaches to signal separation such as the discrete ICA and PCA cannot be applied here as the data cannot be fit into rigid vectors or matrices, which the standard algorithms rely on \cite{hyvarinenIndependentComponentAnalysis2001}. We train our implicit neural ICA model with three components on the spectrogram data points and take a uniform discrete sampling of the learned bases and activations and plot them in \autoref{fig:audio-cqt}. The structure in the original spectrogram is decomposed into the learned bases which are activated across time, where each of the basis functions represents the frequency signature of the corresponding note being played and the source signals show how these bases are activated throughout the audio. We can see minimal leakage between different bases and activations as they are trained to be \emph{maximally independent} through the ICA contrast function.

\section{CONCLUSIONS}
An implicit representation version of PCA and ICA decomposition problems on continuous inputs is presented and a numerical solution framework is developed leveraging implicit neural representations. This general setup allows performing these decompositions on, and extract latent source signals from, irregularly sampled signals where it otherwise would not have been possible. We demonstrate the performance of our algorithm on illustrative examples of various signals and show it achieves decompositions with the desired statistical properties as the traditional PCA/ICA where they are applicable. The ability to process data in this manner can have multiple applications in cases with missing data, irregular sampling (e.g. LIDAR), and in decomposing continuous signals that cannot be fully captured.


\bibliographystyle{IEEEbib}\label{sec:ref}
{\bibliography{refs}}

\begin{thebibliography}{10}

\bibitem{shlensTutorialPrincipalComponent2014}
Jonathon Shlens,
\newblock ``A {{Tutorial}} on {{Principal Component Analysis}},'' Apr. 2014.

\bibitem{roweisEMAlgorithmsPCA1997}
Sam~T Roweis,
\newblock ``{{EM Algorithms}} for {{PCA}} and {{SPCA}},''
\newblock in {\em Advances in {{Neural Information Processing Systems}}}. 1997, vol.~10, MIT Press.

\bibitem{hyvarinenIndependentComponentAnalysis2001}
Aapo Hyvarinen, Juha Karhunen, and Erkki Oja,
\newblock {\em Independent Component Analysis},
\newblock J. Wiley, New York, 2001.

\bibitem{roweisUnifyingReviewLinear1999}
Sam Roweis and Zoubin Ghahramani,
\newblock ``A {{Unifying Review}} of {{Linear Gaussian Models}},''
\newblock {\em Neural Computation}, vol. 11, no. 2, pp. 305--345, Feb. 1999.

\bibitem{amariNewLearningAlgorithm1995}
Shun-ichi Amari, Andrzej Cichocki, and Howard~Hua Yang,
\newblock ``A {{New Learning Algorithm}} for {{Blind Signal Separation}},''
\newblock {\em Advances in Neural Information Processing Systems}, vol. 8, 1995.

\bibitem{cardosoSourceSeparationUsing1989}
J.-F. Cardoso,
\newblock ``Source separation using higher order moments,''
\newblock in {\em International {{Conference}} on {{Acoustics}}, {{Speech}}, and {{Signal Processing}}}, Glasgow, UK, May 1989, IEEE.

\bibitem{schellNonlinearIndependentComponent2023}
Alexander Schell and Harald Oberhauser,
\newblock ``Nonlinear independent component analysis for discrete-time and continuous-time signals,''
\newblock {\em The Annals of Statistics}, vol. 51, no. 2, Apr. 2023.

\bibitem{dauxoisAsymptoticTheoryPrincipal1982}
J.~Dauxois, A.~Pousse, and Y.~Romain,
\newblock ``Asymptotic theory for the principal component analysis of a vector random function: {{Some}} applications to statistical inference,''
\newblock {\em Journal of Multivariate Analysis}, vol. 12, no. 1, pp. 136--154, Mar. 1982.

\bibitem{yao2005functional}
Fang Yao, Hans-Georg M{\"u}ller, and Jane-Ling Wang,
\newblock ``Functional data analysis for sparse longitudinal data,''
\newblock {\em Journal of the American statistical association}, vol. 100, no. 470, pp. 577--590, 2005.

\bibitem{zhong2023nonlinear}
Rou Zhong, Chunming Zhang, and Jingxiao Zhang,
\newblock ``Nonlinear functional principal component analysis using neural networks,''
\newblock {\em arXiv preprint arXiv:2306.14388}, 2023.

\bibitem{loeve1946functions}
Michel Lo{\`e}ve,
\newblock ``Random functions with orthogonal exponential decomposition,''
\newblock {\em The Scientific Review}, vol. 84, pp. 159--162, 1946.

\bibitem{Levy2008}
Bernard~C. Levy,
\newblock {\em Karhunen Loeve Expansion of Gaussian Processes}, pp. 1--47,
\newblock Springer US, Boston, MA, 2008.

\bibitem{tancikFourierFeaturesLet2020}
Matthew Tancik, Pratul~P. Srinivasan, Ben Mildenhall, Sara {Fridovich-Keil}, Nithin Raghavan, Utkarsh Singhal, Ravi Ramamoorthi, Jonathan~T. Barron, and Ren Ng,
\newblock ``Fourier {{Features Let Networks Learn High Frequency Functions}} in {{Low Dimensional Domains}},'' June 2020.

\bibitem{sitzmann2020implicit}
Vincent Sitzmann, Julien Martel, Alexander Bergman, David Lindell, and Gordon Wetzstein,
\newblock ``Implicit neural representations with periodic activation functions,''
\newblock {\em Advances in neural information processing systems}, vol. 33, pp. 7462--7473, 2020.

\bibitem{10.1145/3503250}
Ben Mildenhall, Pratul~P. Srinivasan, Matthew Tancik, Jonathan~T. Barron, Ravi Ramamoorthi, and Ren Ng,
\newblock ``Nerf: representing scenes as neural radiance fields for view synthesis,''
\newblock {\em Commun. ACM}, vol. 65, no. 1, pp. 99–106, Dec. 2021.

\bibitem{subramani2024rethinking}
Krishna Subramani, Paris Smaragdis, Takuya Higuchi, and Mehrez Souden,
\newblock ``Rethinking non-negative matrix factorization with implicit neural representations,''
\newblock {\em arXiv preprint arXiv:2404.04439}, 2024.

\bibitem{subramaniPointCloudAudio2021}
Krishna Subramani and Paris Smaragdis,
\newblock ``Point {{Cloud Audio Processing}},''
\newblock in {\em 2021 {{IEEE Workshop}} on {{Applications}} of {{Signal Processing}} to {{Audio}} and {{Acoustics}} ({{WASPAA}})}, New Paltz, NY, USA, Oct. 2021, pp. 31--35, IEEE.

\bibitem{cichockiRobustLearningAlgorithm1994}
A.~Cichocki, R.~Unbehauen, and E.~Rummert,
\newblock ``Robust learning algorithm for blind separation of signals,''
\newblock {\em Electronics Letters}, vol. 30, no. 17, pp. 1386--1387, Aug. 1994.

\bibitem{shun-ichiamariBlindSourceSeparationsemiparametric1997}
{Shun-Ichi Amari} and J.-F. Cardoso,
\newblock ``Blind source separation-semiparametric statistical approach,''
\newblock {\em IEEE Transactions on Signal Processing}, vol. 45, no. 11, pp. 2692--2700, Nov. 1997.

\bibitem{cbcl}
``{CBCL Face Database \#1, MIT Center For Biological and Computation Learning},'' \url{http://www.ai.mit.edu/projects/cbcl}.

\bibitem{brownCalculationConstantSpectral1991}
Judith~C. Brown,
\newblock ``Calculation of a constant {{{\emph{Q}}}} spectral transform,''
\newblock {\em The Journal of the Acoustical Society of America}, vol. 89, no. 1, pp. 425--434, Jan. 1991.

\bibitem{velascoCONSTRUCTINGINVERTIBLECONSTANTQ2011}
Gino~Angelo Velasco, Nicki Holighaus, Monika D{\"o}rfler, and Thomas Grill,
\newblock ``{{Constructing an invertible constant-Q transform with nonstationary Gabor frames}},''
\newblock in {\em Proceedings of {{DAFX11}}}, Paris, 2011, vol.~33.

\end{thebibliography}

\end{document}